\documentclass[letterpaper]{article} 
\usepackage[]{aaai24}  
\usepackage{times}  
\usepackage{helvet}  
\usepackage{courier}  
\usepackage[hyphens]{url}  
\usepackage{graphicx} 
\usepackage{natbib}  
\usepackage{caption} 
\usepackage{algorithm}
\usepackage{algorithmicx}
\usepackage[noend]{algpseudocode}
\usepackage{mathtools}  
\newtheorem{principle}{Principle}
\usepackage{tikz}

\usepackage{newfloat}
\usepackage{listings}
\DeclareCaptionStyle{ruled}{labelfont=normalfont,labelsep=colon,strut=off} 
\floatstyle{ruled}
\newfloat{listing}{tb}{lst}{}
\floatname{listing}{Listing}

\title{Unsupervised Bias Detection in College Student Newspapers}
\author{
    Adam Lehavi\textsuperscript{\rm 1}\thanks{Equal contribution},
    William McCormack\textsuperscript{\rm 1}\footnotemark[1],
    Noah Kornfeld\textsuperscript{\rm 1},
    Solomon Glazer\textsuperscript{\rm 1}
}
\affiliations{
    \textsuperscript{\rm 1}Social Media Climate Initiative\\
    alehavi@usc.edu, willmccormack@g.ucla.edu, nekornfeld@wisc.edu, solomon.glazer@baruchmail.cuny.edu
}

\begin{document}

\maketitle

\begin{abstract}
This paper presents a pipeline with minimal human influence for scraping and detecting bias on college newspaper archives. This paper introduces a framework for scraping complex archive sites that automated tools fail to grab data from, and subsequently generates a dataset of 14 student papers with 23,154 entries. This data can also then be queried by keyword to calculate bias by comparing the sentiment of a large language model summary to the original article. The advantages of this approach are that it is less comparative than reconstruction bias and requires less labelled data than generating keyword sentiment. Results are calculated on politically charged words as well as control words to show how conclusions can be drawn. The complete method facilitates the extraction of nuanced insights with minimal assumptions and categorizations, paving the way for a more objective understanding of bias within student newspaper sources.
\end{abstract}

\section{Introduction}

In a world filled with so much information, being able to automatically get and understand data can save countless hours. Across various fields of study, data scraping and sentiment analysis can enable researchers to reveal nuanced patterns for meaningful insights \cite{byrne2013comparison, gite2021explainable}. Sentiment analysis, or using machine learning to discern text's tone and emotion, can be used for financial and social benefit. In the public health sector, sentiment analysis of real-time tweets was shown to detect and localize Covid-19 outbreaks \cite{alamoodi2021sentiment}.

In the field of media study, detecting media bias has broad implications in understanding what source to trust \cite{anzum2022biases}. Being able to do so in an automated and streamlined manner could allow for both the monitoring and ranking of existing media as well as scoring for future content to improve its credibility.

Despite the noteworthy progress made in these disciplines, several obstacles remain. A considerable portion of automated extraction methods grapple with cases of multi-tiered content extraction, particularly when from voluminous media archives \cite{bhatt2011multimedia}. Simultaneously, past bias identification strategies frequently depend on labeled and grouped data to either highlight how media outlets differ from one another, or what outlets’ stances are on subjects. This inherently pushes the data's skews and assumptions onto the results, which can be avoided by looking at media outlets as a population to draw conclusions from \cite{budak2016fair}.

To tackle these challenges, we propose an innovative, largely automated pipeline for unsupervised sentiment analysis. This new approach aims to detect media bias with fewer assumptions and less classification, thus promoting a more nuanced and precise understanding of media bias. By presenting a unique approach that combines data extraction with unsupervised sentiment analysis, this research seeks to contribute to ongoing scholarly conversations about media bias and its detection \cite{montoyo2012subjectivity}. In the sections that follow, we delve deeper into the methodology and findings of this research, underscoring its potential for enhancing current approaches to media bias detection and analysis.

In this research, we introduce a novel methodology that merges state-of-the-art data extraction and unsupervised sentiment analysis, contributing significantly to the ongoing academic discourse on media bias and its detection. We start by positioning our work within the swiftly evolving realms of sentiment detection and bias identification, and underscore how our approach extends the existing body of knowledge. We proceed to detail our workflow which includes data extraction from college newspaper archives, followed by an overview of our sentiment analysis pipeline. The subsequent sections focus on the examination of our methods, a comparative study of statistical outcomes across various text granularities, an acknowledgement of potential constraints of our approach, and a forecast of future research directions building upon this foundation.

\section{Literature Review}
\subsection{Data Scraping}
Data scraping has emerged as a popular and efficient method for mass data collection, offering numerous benefits such as ease of use. Researchers can now employ unsupervised models to rapidly extract, synthesize, and organize vast amounts of data without human oversight. This technological advancement has revolutionized the research landscape, enabling studies involving thousands of articles across various websites in a time-efficient manner, a feat that would otherwise be nearly impossible.

Numerous approaches to data scraping have been developed over the years. One of the earliest techniques, pioneered in 1997 \cite{hammer1997extracting}, involved the manual coding of multiple extraction programs, each tailored to extract data from individual websites based on their observed format and patterns. Another approach, introduced in 2001 \cite{chang2001iepad}, focused on uncovering patterns within hidden databases that underpin visible web pages. However, both of these methods were unsuitable for our study, as final content pages had repetitive HTML but each archive's layout presented a unique challenge, with sites often having flaws and inconsistencies.

A third approach, introduced by Liu, Grossman, and Zhai in 2003 and known as MDR \cite{liu2003mining}, offered greater relevance to our final methodology. This technique leveraged the similarities found in the HTML structures of article pages to design a program capable of extracting data based on these HTML signatures. While the differences in HTML signatures across different student publications prevented us from entirely generalizing one program to extract data, we show in our pipeline a partial implementation in the form of grouped types of websites.

A modern approach demonstrated by Kusumasari and Prabowo in 2020 \cite{kusumasari2020scraping} utilized parameters such as specific keywords and time periods to scrape data from Twitter for trends. Although many approaches similar to this exist for social media scraping, our college media was flawed in that it lacked consistent format, easy searching, and the many other utilities of organized standardized media content.

\subsection{College Media Data Collection}
Existing research in college media bias has primarily focused on political party views regarding election influence. Aimee Burch and Raluca Cozma's 2016 study \cite{burch2016student} investigated how student publications in swing states covered the 2012 presidential election, concluding that these newspapers displayed a generally more neutral tone compared to their professional counterparts. Similarly, Hans Schmidt's 2015 study \cite{schmidt2015student} aimed to assess whether student journalists' personal preferences and biases influenced their article content during the 2008 and 2012 presidential elections. 

Both of these method suffered from the pitfalls of manual data collection and organization, and drew conclusions from limited data with hand-chosen papers. Leveraging a machine-based collection method significantly reduces human resource requirements in scraping, organizing, and summarizing. Using unsupervised bias, as well, one can focus on distance from a ground truth, without an associated need to draw party lines.

\subsection{Bias Detection}
Sentiment analysis plays a crucial role in bias detection, as machine learning models can effectively classify text into positive, negative, or neutral attitudes. This attribute has been extensively researched, particularly in the context of identifying biases in political news articles. A study by Minh Vu in 2017 \cite{vu2017political} investigated whether politically charged information consumed by individuals is influenced by their pre-existing political opinions. The study classified articles based on liberal, conservative, or neutral attitudes, comparing these classifications to the sentiment model's analysis of each article's sentences. While the results were promising, the study employed a relatively simplistic approach to sentiment analysis, solely categorizing each article with one political position and exclusively based on analysis of each individual sentence.

Furthermore, a study published in 2020 \cite{mozafari2020hate} explored the use of sentiment analysis, particularly with the BERT natural language processing model, to detect hate speech in tweets and categorize them into different types of hate speech, such as racism and sexism. Similar to the previous study, this research employed a classification strategy and labeled tweets as either containing hate speech or not.

In our approach, each article is classified on a spectrum of positive, negative, and neutral biases concerning multiple topics. Moreover, our methodology allowed for the separation of article biases based on overall sentiment, paragraph-based sentiment, and sentence-based sentiment, providing a more comprehensive analysis compared to the previous studies' binary classifications.

\subsection{Text Summarization Applications}
Text summarization, a powerful tool for streamlining the data collection process, involves programming AI models to coherently and concisely summarize large volumes of text. In systematic reviews, machine-based text summarization proves exceptionally useful during the initial stages. For instance, text mining and summarizing can be used to find relevant studies based on flagged words \cite{thomas2011applications}. Our study builds upon the utility of text summarization, using summarization to establish a ground truth of what the basis of a newspaper article is.

\section{Methodology}

\begin{figure*}[ht]
\centering
\includegraphics[width=0.9\linewidth]{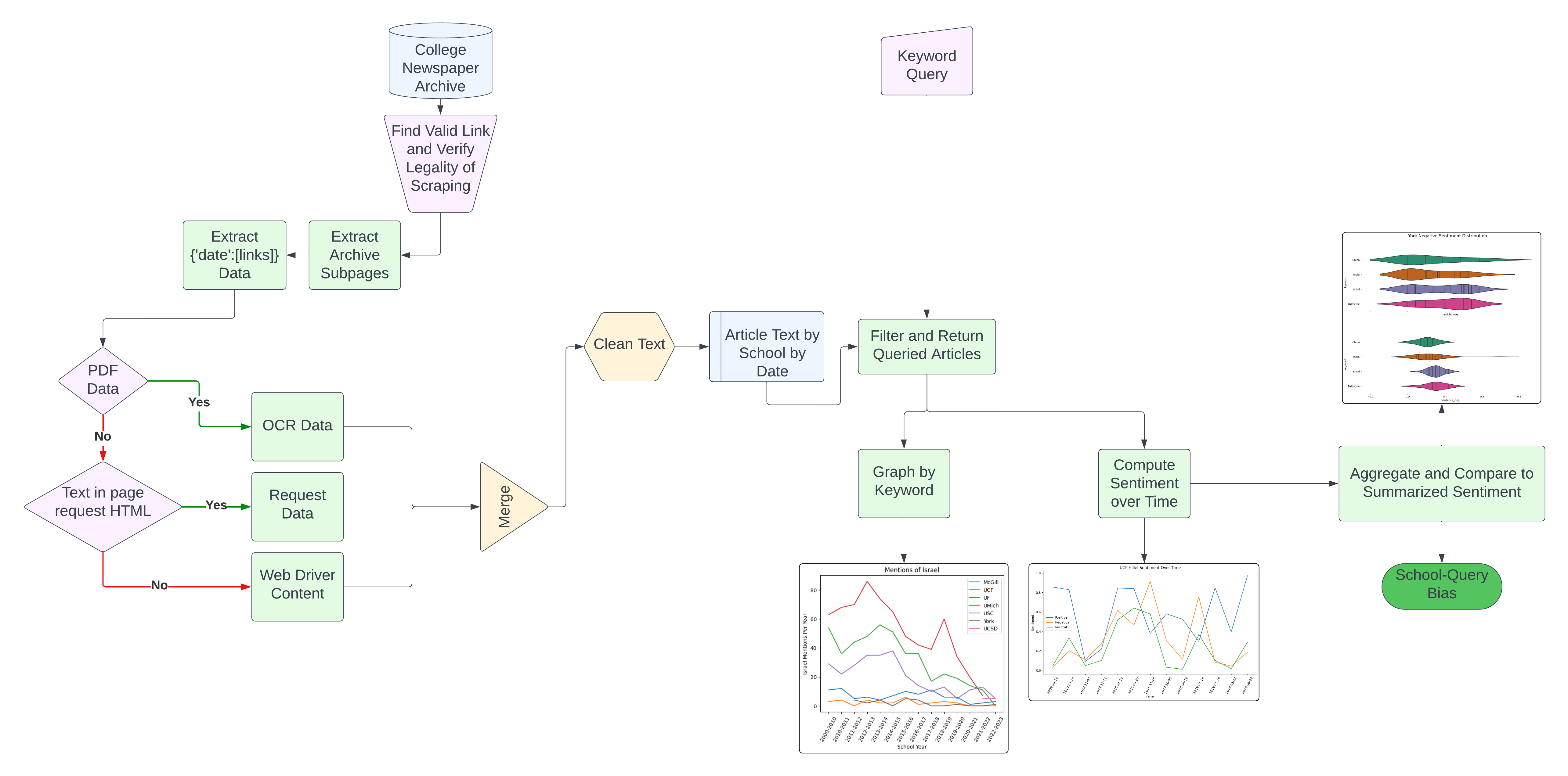} 
\caption{Workflow Layout}
\label{fig:workflow_dia}
\end{figure*}

Our complete workflow is shown in figure \ref{fig:workflow_dia}, with all code and intricacies described relating back to this general layout.

\subsection{Inspecting for Validity}

In our research, we focused on student newspaper archives from 16 public and private universities across the United States and Canada. Schools chosen were chosen from Hillel's list of 60 largest schools by Jewish student population \cite{hillel2022}. This is because schools with larger Jewish student populations are more likely to have more articles relating to Israel and Palestine both in a worldwide sense and referring to campus events. Without this precaution, campus events would be more grouped with queries of India and China and introduce a fault in results.

Of the articles on this list, schools were not explored sequential to ranking. Rather, schools were chosen to generate a mix of types of schools in results, by location, size, and private or public. School media chosen was chosen by searching "[School] Student Newspaper" followed by "[Student Newspaper] Archive", invalidating cases where schools did not have any large media affiliated or unaffiliated with campuses. Archives were not pursued further if the organizations had messages prohibiting scraping, or if no archive could be found. Archives going back to 2009 were preferable for the sake of consistency. This process and a constraint of checking 30 schools because of time caps led us to the schools we have.

\subsection{Extracting Archive Subpages and Article Pages}

The biggest block to complete automation is taking the first page of an archive to extrapolate all archive subpages and all article pages resulting from them. The archive subpages can be accessed using one of the following methods: by examining the page content to identify the maximum page number in a bottom navigation bar, by inspecting the original subpage link to manually navigate to future pages, by going to the second subpage to obtain this link information, or by analyzing network data to directly access the backend API. For professional sites, a sitemap could generate all this information in a clean manner. However, many of these sites were built with old technologies. 

Once on these subpages, something similar to MDR could be considered for getting all links for given dates. However, the content tags were volatile enough such that it was faster to manually grab the needed tags, and then use a script to iterate over information for day-by-day dictionary creation.

\subsection{Scraping Content}

All archives can be grouped as having recurring PDFs, having text data directly on the site, or loading in text data with JavaScript. PDFs generated particularly noisy content, even after processing. As such, the gathered data was too fragmented to safely deduce bias, and so these schools are not included in final results. These schools are the motivating factor for why data is grouped by date, and the hope is that future improvements in provessing can lead to their inclusion. Text data from the other sites was sufficient, with the only notable flaw being watermarks. Certain schools always generated headers or footers, either related to the web content itself, or copyright notices. These can be trimmed off in processing steps, but are maintained in data gathering such that the data itself is a reflective of the original source as possible.

\subsection{Querying and Processing}

Querying is done by a simple keyword search. Experiments with modifying the regular expression to search for instances of the word where there was a portion cut off were used, but showed little additional yield for the added runtime. The main result of this simple method is the inclusion of articles where the keyword is not the topic. 

As has been done in past literature, the main topic could be extracted through a large language model and used for filtering. This method was avoided as the conclusions it would generate would be heavily reliant on the nature of the language model used and the particular hyperparameters applied.

\subsection{Sentiment and Summarization}

Sentiment is computed through the use of either NLTK \cite{bird2009natural}, a leading python natural language library, or through the BART model \cite{lewis2019bart} from the HuggingFace library. 

Results shown use NLTK's VADER model for sentiment intensity. While a larger, more sophisticated model better captures the nuance of the text, the runtime for doing so on a large corpus of text makes it generally impractical. Beyond this, a simple sentiment model suffices.

Text is calculated for a summary of the entire article, for paragraph-by-paragraph summaries, and for each sentence of the original article. In all of these groupings, the sentiment model needs only calculate sentiment a single sentence at a time, to be averaged and grouped. This is to focus on the language used, and so allows for basic models to still generate meaningful conclusions.

Summarization is done using Google's T5 model \cite{raffel2020exploring}. The particular summarizer, similar to the choice for sentiment, is not the focus of the work and should not heavily hinder output. This model, in particular, was chosen because it shows state-of-the-art performance on many tasks and was trained on broad datasets, making it less likely to lose information when important.

\subsection{Bias Calculation}

\begin{principle}
\label{thm:bias_from_sent}
    Media bias can be measured as the difference between what a media outlet believes the truth to be to what they report.
\end{principle}

We calculate bias under principle \ref{thm:bias_from_sent}. The most central difference between this and what other works have done is that there is no single truth of an event. The concept behind this is to encapsulate that media should aim to report facts, even if the reporting of certain facts with certain nuances leads to a change in belief. Imbuing emotion and drawing conclusions in an article is a focal and preventative source of bias.

As such, sentiment is found on article, paragraph, and sentence level granularities. Article-wide sentiment is found by summarizing the entire article and calculating sentiment on the summary. Paragraph sentiment is done by summarizing each paragraph and averaging sentiment. Sentence sentiment is the average from each sentence. When we average, we keep negative, neutral, and positive separate from one another.

Bias is calculated as the difference between the article summary sentiment, considered the media's truth, and the sentence summary, considered what an outlet reports.

\subsection{Conclusions to be Drawn}

Based on the nuances of the process, it is worth explicitly noting that a conclusion must contextualize the methodology.
\begin{itemize}
    \item \textbf{Valid Conclusion:} A given school's articles show more bias when including a certain topic.
    \item \textbf{Valid Conclusion:} When looking at the population of schools or articles of a given school, one keyword seems to have more biased reporting in articles with it compared to other keywords.
    \item \textbf{Invalid Conclusion:} A given school is biased in favor or against a certain topic.
\end{itemize}

Conclusions that certain schools are more biased than others in reporting are plausible, and compared to reconstruction bias carry more weight as they are more independent of political ideology of the population. However, without isolating variables like the topics of articles, weekly or recurring reporting segments, or which authors contribute, broad conclusions are ill-mannered.

\section{Results}
All code was run on Google Colab notebooks with the default Intel CPU and 13GB of RAM, or on local desktop computers with equal or lesser RAM and no GPUs. Running sentiment and summarization for all shown keywords took roughly 40 hours, with the majority of time spent on summarization.

\subsection{Scraping}
In total, 14 student papers with 23,154 entries were collected and aggregated, as shown in table \ref{tab:school_art_count}. This was done using the pipeline methodology. This means that schools listed were manually guided towards constructing a dictionary of dates and associated article links. Past that point, all further steps were automated. Certain schools, UF for example, had issues with encoded text, and so could not be used for sentiment.
\begin{table*}[ht]
\centering
\caption{Data is from school student newspapers, not all of which are affiliated with the schools. We do not claim this data nor the opinions expressed in it to be representative of the schools listed. Conclusions of bias are for the sake of research and not to condemn any of the associated mentioned press or their opinions. Earliest and latest date in the table refer to those in the database, and not inherently those in the archive. Count is grouped by day, and not by article.}
\label{tab:school_art_count}
\begin{tabular}{||r | r | r | r | r | r||} 
\hline
School Name & Student Newspaper & source & Earliest Date & Latest Date & Count\\ 
\hline \hline
AU & The Eagle & txt & 2009/02/12 & 2023/05/06 & 805\\
\hline
CMU & The Tartan & txt & 2009/01/19 & 2023/05/01 & 346\\
\hline
GW & The Hatchet & txt & 2009/01/12 & 2023/05/25 & 877\\
\hline
Georgetown & The Hoya & txt & 2010/01/16 & 2017/09/13 & 1609\\
\hline
Harvard & Harvard Crimson & txt & 2009/01/02 & 2012/01/02 & 4673\\
\hline
LIU & Seawanhaka & txt & 2010/10/13 & 2023/05/04 & 480\\
\hline
McGill & The McGill Daily & txt & 2009/01/12 & 2023/04/03 & 866\\
\hline
UCF & UCF Today & txt & 2009/07/12 & 2013/05/29 & 3049\\
\hline
UCSD & The Guardian & PDF & 2008/11/13 & 2023/01/30 & 658\\
\hline
UF & The Alligator & PDF & 2009/01/06 & 2015/01/21 & 1804\\
\hline
UMich & Michigan Daily & PDF & 2009/01/07 & 2015/10/13 & 1773\\
\hline
USC & Daily Trojan & txt & 2009/05/20 & 2013/11/30 & 3118\\
\hline
USF & The Oracle & txt & 2009/01/04 & 2014/11/24 & 2668\\
\hline
York & Excalibur & txt & 2010/09/25 & 2023/04/26 & 428\\
\hline
\end{tabular}
\end{table*}

\subsection{Sentiment Distribution}
\begin{figure}[ht]
    \centering
    \includegraphics[width=0.9\linewidth]{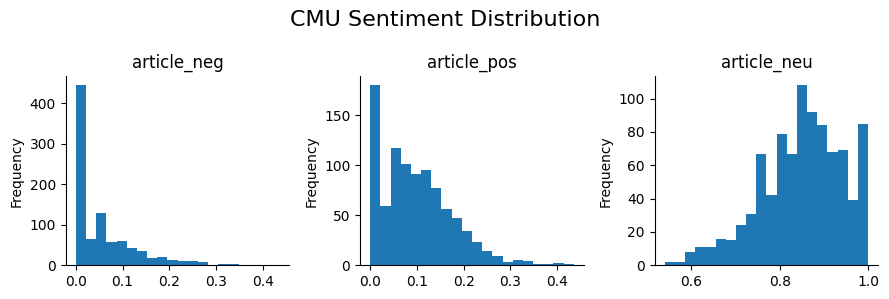}
    \caption{Distribution of Article Sentiment Scores for CMU's Articles for Keywords [India, China, Israel, Palestine]}
    \label{fig:cmu-dist}
\end{figure}

Sentiment distribution both for article and sentence granularity for most schools appeared similar to figure \ref{fig:cmu-dist}. In this sense, positive and negative sentiment defaulted to 0, and neutral sentiment to 1. These graphs inspired a much more heavy focus on negative sentiment for comparison, as there were less false associations than positive sentiment.

\begin{figure}[ht]
    \centering
    \includegraphics[width=0.9\linewidth]{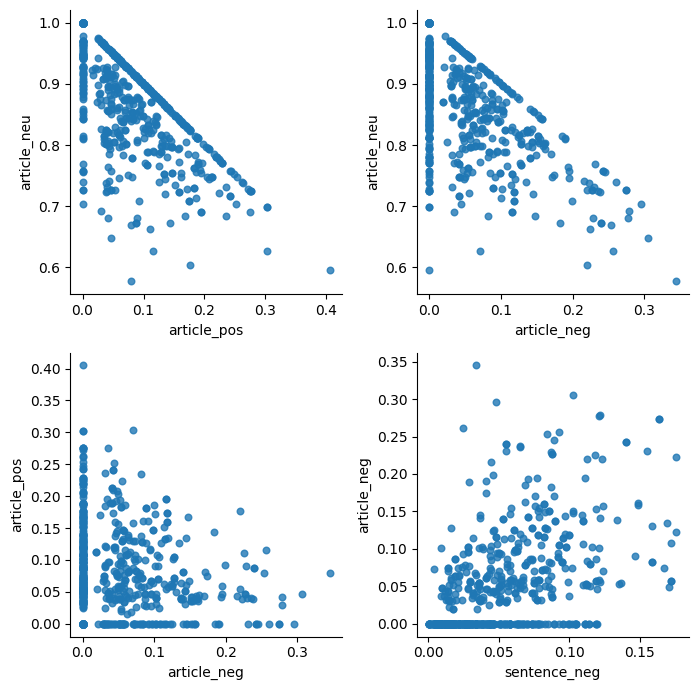}
    \caption{Sentiment-by-sentiment Distribution for Georgetown for Keywords [India, China, Israel, Palestine]}
    \label{fig:george_sent_by_sent}
\end{figure}

When graphing the different sentiment scores relative to each other, such as in figure \ref{fig:george_sent_by_sent}, there were a few general trends observed. Sentence sentiment was much less likely to be zero than article sentiment, which is expected of an averaged value. Matching sentiments over different granularities appeared to trend together, and differing sentiments generally trended against one another, with articles having to be primarily one of the three.

\subsection{Bias}
Bias for schools ran is shown in table \ref{tab:bias_all}. Not all schools were used because of runtime and model constraints. T5 has a context limit of 500 tokens, and so schools that continued to show articles of over 500 tokens of length were avoided for being problematic. In the table, the values shown are all calculated as the percentage point difference from the article to sentence, meaning a positive value of 2.86 indicates the article is 2.86 percentage points higher in whichever category. Certain schools seem to generate more bias than others over almost all cases. York, which gets the largest bias values, also has only 73 articles within the 4 keywords. Georgetown and CMU, both having over 125 articles, are better for comparing high and low bias. Certain keywords bring more associated bias with them, such as Palestine when compared to Israel.
\begin{table*}[ht]
\centering
\caption{Bias Results for All Calculated Schools}
\begin{tabular}{|r|r|r|r|r|r|r|r|r|r|r|r|r|}
\hline
&\multicolumn{3}{c}{Israel}\vline & \multicolumn{3}{c}{Palestine}\vline & \multicolumn{3}{c}{India} \vline& \multicolumn{3}{c}{China}\vline\\ \hline
School\_Name & pos & neg & neu &pos & neg & neu & pos & neg & neu & pos & neg & neu \\
\hline \hline
LIU &1.04 & 0.92 & -1.91 & -1.70 & -1.00 & \textbf{2.80} & -1.11 & \textbf{2.16} & -1.04 & -2.86 & \textbf{1.74} & 1.11  \\ \hline
Georgetown &-1.48 & \textbf{-1.07} & \textbf{2.56} & -2.42 & -0.29 & 2.71 & -2.78 & 0.26 & 2.53 & -2.08 & -0.27 & 2.35 \\ \hline
CMU &0.05 & -0.25 & 0.20 & -0.67 & 1.69 & -1.02 & -1.38 & -0.65 & 2.03 & -1.22 & -0.92 & 2.15  \\ \hline
AU &-1.61 & 0.96 & 0.66 & 0.31 & 2.01 & -2.31 & -1.05 & -1.03 & 2.06 & -3.70 & 0.15 & \textbf{3.56} \\ \hline
USC& &  &  &  &  &  &  &  &  & -3.36 & -0.02 & 3.39\\ \hline
York&\textbf{-1.82} & 1.06 & 0.76 & \textbf{-4.69} & \textbf{2.60} & 2.10 & \textbf{-3.76} & 0.46 & \textbf{3.30} & \textbf{-3.71} & 1.58 & 2.09 \\
\hline
\end{tabular}
\label{tab:bias_all}
\end{table*}

For American University, additional keywords were calculated, as displayed in table \ref{tab:bias_au}. Within similar types of keywords, such as countries or political descriptions, conclusions can be drawn on the basis of absolute scale. However, there are many words or cases where a keyword has more bias in a certain category but not others. These examples are difficult to decipher as of now.

\begin{table}[ht]
\centering
\caption{Sentiment for certain keywords for AU}
\begin{tabular}{||r|r|r|r||}
\hline
&pos & neg & neu\\
\hline \hline
Israel & -1.61 & 0.96 & 0.66 \\ \hline 
Palestine & 0.31 & 2.01 & -2.31 \\ \hline 
India & -1.05 & -1.03 & 2.06 \\ \hline 
China & -3.70 & 0.15 & 3.56\\ \hline 
Conservative & -2.37 & -0.65 & 3.03 \\ \hline
Democrat & -1.28 &	-0.13	&1.40 \\ \hline
Liberal & -3.49 &	-0.35 &	3.84 \\ \hline
Republican & -2.77 &	1.54&	1.21 \\ \hline
\end{tabular}
\label{tab:bias_au}
\end{table}

\begin{figure}[ht]
    \centering
    \includegraphics[width=0.9\linewidth]{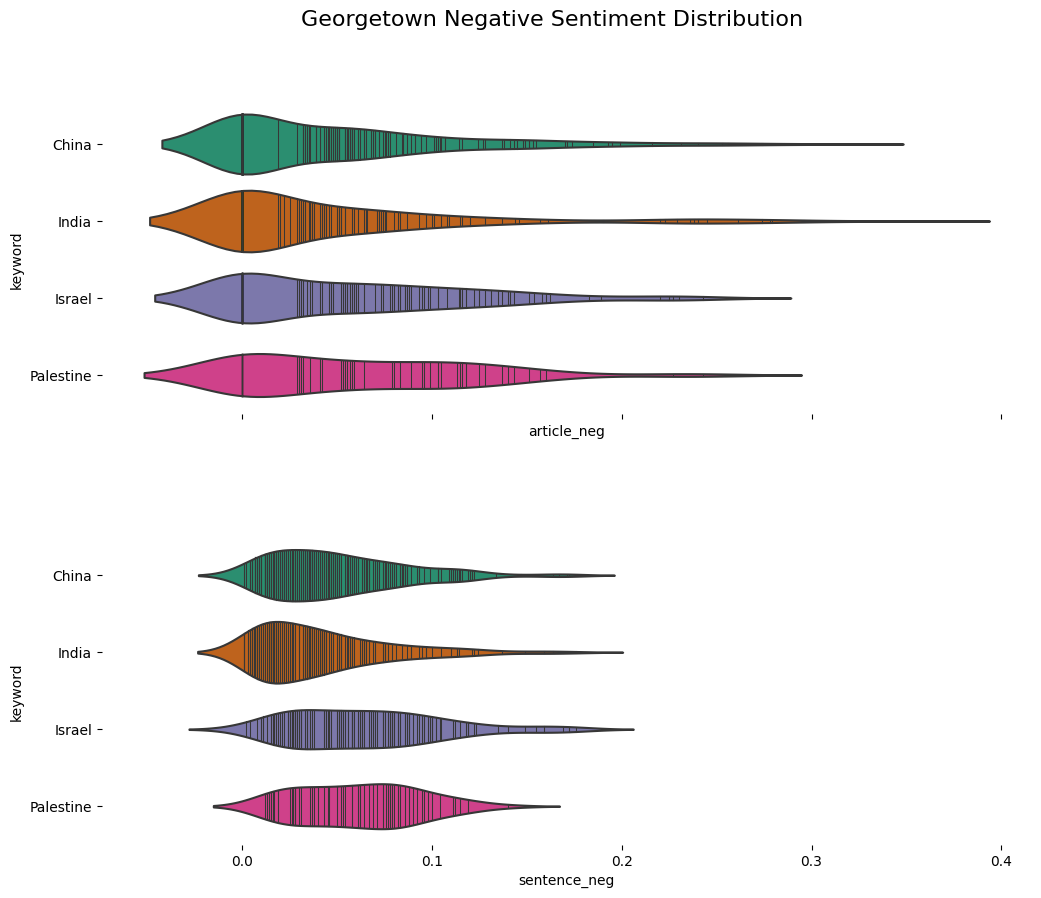}
    \caption{Negative sentiment distribution for Georgetown on a violin plot. Although the axis extends beyond 0, no negative values are present or allowable.}
    \label{fig:george-violin}
\end{figure}

\begin{figure}[ht]
    \centering
    \includegraphics[width=0.9\linewidth]{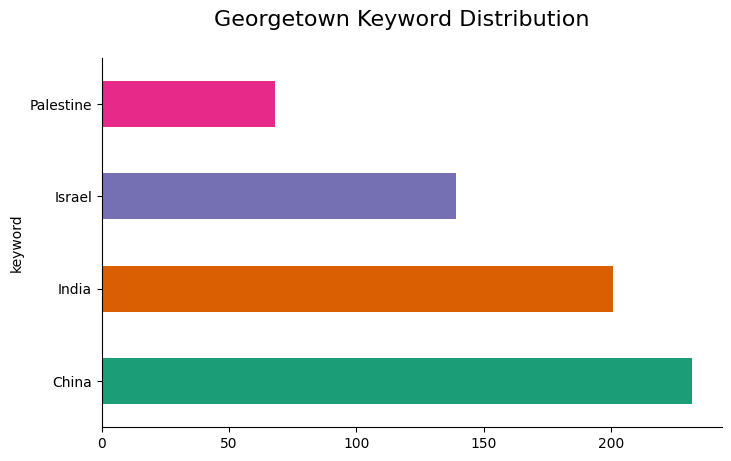}
    \caption{Keyword distribution for Georgetown, where certain articles can appear for multiple keywords}
    \label{fig:george-key-dist}
\end{figure}

The bias of given keywords can also be compared with violin plots or box and whisker plots of the sentiment distribution to get a fuller picture of what the raw values were and how they may have been skewed. An example is shown in figure \ref{fig:george-violin} for Georgetown's negative bias, which can be furthered contextualized with keyword distribution from figure \ref{fig:george-key-dist}.

\section{Conclusion}

In this study, we introduce and use a complete pipeline for detecting and comparing bias in student newspaper archives. Conclusions about the general group of schools or any one in particular are difficult to make. Rather, we generate exploratory results. With these results, a school can focus on balancing reporting between keywords, or curating future articles to match the balance. Compared to prior research, we succeed in generating our measure of bias unsupervised. We scrape college newspapers while overcoming many of the challenges in site variance and noise. 

Future work can entail more in-depth comparison of summarization and sentiment models. More advanced processing and computation could allow for a more complete bias comparison over all scraped schools. The workflow can also hopefully be further automated to expand the dataset. Comparing more schools or even comparing school media to professional media could base scores of bias more soundly.

All code and data used to generate results is present at https://anonymous.4open.science/r/UnsupervisedBias-C0F0. We hope that the field of college media bias detection can grow to acknowledge and explore more of the nuances in successfully generated automated results.

\bibliography{aaai24}

\begin{thebibliography}{20}
\providecommand{\natexlab}[1]{#1}

\bibitem[{Alamoodi et~al.(2021)Alamoodi, Zaidan, Zaidan, Albahri, Mohammed,
  Malik, Almahdi, Chyad, Tareq, Albahri et~al.}]{alamoodi2021sentiment}
Alamoodi, A.~H.; Zaidan, B.~B.; Zaidan, A.~A.; Albahri, O.~S.; Mohammed, K.~I.;
  Malik, R.~Q.; Almahdi, E.~M.; Chyad, M.~A.; Tareq, Z.; Albahri, A.~S.; et~al.
  2021.
\newblock Sentiment analysis and its applications in fighting COVID-19 and
  infectious diseases: A systematic review.
\newblock \emph{Expert systems with applications}, 167: 114155.

\bibitem[{Anzum, Asha, and Gavrilova(2022)}]{anzum2022biases}
Anzum, F.; Asha, A.~Z.; and Gavrilova, M.~L. 2022.
\newblock Biases, fairness, and implications of using AI in social media data
  mining.
\newblock In \emph{2022 International Conference on Cyberworlds (CW)},
  251--254. IEEE.

\bibitem[{Bhatt and Kankanhalli(2011)}]{bhatt2011multimedia}
Bhatt, C.~A.; and Kankanhalli, M.~S. 2011.
\newblock Multimedia data mining: state of the art and challenges.
\newblock \emph{Multimedia Tools and Applications}, 51: 35--76.

\bibitem[{Bird, Klein, and Loper(2009)}]{bird2009natural}
Bird, S.; Klein, E.; and Loper, E. 2009.
\newblock \emph{Natural language processing with Python: analyzing text with
  the natural language toolkit}.
\newblock " O'Reilly Media, Inc.".

\bibitem[{Budak, Goel, and Rao(2016)}]{budak2016fair}
Budak, C.; Goel, S.; and Rao, J.~M. 2016.
\newblock Fair and balanced? Quantifying media bias through crowdsourced
  content analysis.
\newblock \emph{Public Opinion Quarterly}, 80(S1): 250--271.

\bibitem[{Burch and Cozma(2016)}]{burch2016student}
Burch, A.; and Cozma, R. 2016.
\newblock Student election stories use more diverse news sources.
\newblock \emph{Newspaper Research Journal}, 37(3): 235--248.

\bibitem[{Byrne, Jordan, and Welle(2013)}]{byrne2013comparison}
Byrne, M.~D.; Jordan, T.; and Welle, T. 2013.
\newblock Comparison of manual versus automated data collection method for an
  evidence-based nursing practice study.
\newblock \emph{Applied Clinical Informatics}, 4(01): 61--74.

\bibitem[{Chang and Lui(2001)}]{chang2001iepad}
Chang, C.-H.; and Lui, S.-C. 2001.
\newblock IEPAD: Information extraction based on pattern discovery.
\newblock In \emph{Proceedings of the 10th international conference on World
  Wide Web}, 681--688.

\bibitem[{Gite et~al.(2021)Gite, Khatavkar, Kotecha, Srivastava, Maheshwari,
  and Pandey}]{gite2021explainable}
Gite, S.; Khatavkar, H.; Kotecha, K.; Srivastava, S.; Maheshwari, P.; and
  Pandey, N. 2021.
\newblock Explainable stock prices prediction from financial news articles
  using sentiment analysis.
\newblock \emph{PeerJ Computer Science}, 7: e340.

\bibitem[{Hammer et~al.(1997)Hammer, Garcia-Molina, Cho, Aranha, and
  Crespo}]{hammer1997extracting}
Hammer, J.; Garcia-Molina, H.; Cho, J.; Aranha, R.; and Crespo, A. 1997.
\newblock Extracting semistructured information from the Web.
\newblock In \emph{Proceedings of the workshop on management of semistructured
  data}, volume~10. Tucson, Arizona: ACM.

\bibitem[{Hillel(2022)}]{hillel2022}
Hillel. 2022.
\newblock Top 60 Jewish Colleges.

\bibitem[{Kusumasari and Prabowo(2020)}]{kusumasari2020scraping}
Kusumasari, B.; and Prabowo, N. P.~A. 2020.
\newblock Scraping social media data for disaster communication: how the
  pattern of Twitter users affects disasters in Asia and the Pacific.
\newblock \emph{Natural Hazards}, 103(3): 3415--3435.

\bibitem[{Lewis et~al.(2019)Lewis, Liu, Goyal, Ghazvininejad, Mohamed, Levy,
  Stoyanov, and Zettlemoyer}]{lewis2019bart}
Lewis, M.; Liu, Y.; Goyal, N.; Ghazvininejad, M.; Mohamed, A.; Levy, O.;
  Stoyanov, V.; and Zettlemoyer, L. 2019.
\newblock Bart: Denoising sequence-to-sequence pre-training for natural
  language generation, translation, and comprehension.
\newblock \emph{arXiv preprint arXiv:1910.13461}.

\bibitem[{Liu, Grossman, and Zhai(2003)}]{liu2003mining}
Liu, B.; Grossman, R.; and Zhai, Y. 2003.
\newblock Mining data records in web pages.
\newblock In \emph{Proceedings of the ninth ACM SIGKDD international conference
  on Knowledge discovery and data mining}, 601--606.

\bibitem[{Montoyo, Mart{\'\i}nez-Barco, and
  Balahur(2012)}]{montoyo2012subjectivity}
Montoyo, A.; Mart{\'\i}nez-Barco, P.; and Balahur, A. 2012.
\newblock Subjectivity and sentiment analysis: An overview of the current state
  of the area and envisaged developments.
\newblock \emph{Decision Support Systems}, 53(4): 675--679.

\bibitem[{Mozafari, Farahbakhsh, and Crespi(2020)}]{mozafari2020hate}
Mozafari, M.; Farahbakhsh, R.; and Crespi, N. 2020.
\newblock Hate speech detection and racial bias mitigation in social media
  based on BERT model.
\newblock \emph{PloS one}, 15(8): e0237861.

\bibitem[{Raffel et~al.(2020)Raffel, Shazeer, Roberts, Lee, Narang, Matena,
  Zhou, Li, and Liu}]{raffel2020exploring}
Raffel, C.; Shazeer, N.; Roberts, A.; Lee, K.; Narang, S.; Matena, M.; Zhou,
  Y.; Li, W.; and Liu, P.~J. 2020.
\newblock Exploring the limits of transfer learning with a unified text-to-text
  transformer.
\newblock \emph{The Journal of Machine Learning Research}, 21(1): 5485--5551.

\bibitem[{Schmidt(2015)}]{schmidt2015student}
Schmidt, H.~C. 2015.
\newblock Student newspapers show opinion article political bias.
\newblock \emph{Newspaper Research Journal}, 36(1): 6--23.

\bibitem[{Thomas, McNaught, and Ananiadou(2011)}]{thomas2011applications}
Thomas, J.; McNaught, J.; and Ananiadou, S. 2011.
\newblock Applications of text mining within systematic reviews.
\newblock \emph{Research synthesis methods}, 2(1): 1--14.

\bibitem[{Vu(2017)}]{vu2017political}
Vu, M. 2017.
\newblock Political news bias detection using machine learning.
\newblock \emph{URL: https://pdfs. semanticscholar.
  org/8445/2eb068bdfe7d5809734a5da8f5c7d10bebfa. pdf}.

\end{thebibliography}

\end{document}